# Human Activity Recognition Using Multichannel Convolutional Neural Network

Niloy Sikder
*Computer Science & Engineering Discipline*
*Khulna University*
Khulna, Bangladesh
niloysikder333@gmail.com

Abu Shamim Mohammad Arif
*Computer Science & Engineering Discipline*
*Khulna University*
Khulna, Bangladesh
shamimarif@yahoo.com

Md. Sanaullah Chowdhury
*Electronics & Communication Engineering Discipline*
*Khulna University*
Khulna, Bangladesh sanaullahashfat@gmail.com

Abdullah-Al Nahid
*Electronics & Communication Engineering Discipline*
*Khulna University*
Khulna, Bangladesh
nahid.ece.ku@gmail.com

*Abstract*—Human Activity Recognition (HAR) simply refers to the capacity of a machine to perceive human actions. HAR is a prominent application of advanced Machine Learning and Artificial Intelligence techniques that utilize computer vision to understand the semantic meanings of heterogeneous human actions. This paper describes a supervised learning method that can distinguish human actions based on data collected from practical human movements. The primary challenge while working with HAR is to overcome the difficulties that come with the cyclostationary nature of the activity signals. This study proposes a HAR classification model based on a two-channel Convolutional Neural Network (CNN) that makes use of the frequency and power features of the collected human action signals. The model was tested on the UCI HAR dataset, which resulted in a 95.25% classification accuracy. This approach will help to conduct further researches on the recognition of human activities based on their biomedical signals.

*Keywords—HAR, human action recognition, human activity classification, multichannel CNN, UCI HAR, feature extraction*

## I. Introduction

We, as humans, possess an amazing skill to comprehend information that others pass on through their movements like the gesture of a certain body part or the motion of the entire body. We can differentiate among human postures, track complex human motions, and evaluate human-object interactions to realize what they are doing, and even deduce what they intend to do. Even though these are advanced recognition functionalities performed by the brain based on the images of the surroundings captured by the eyes, the process occurs almost autonomously to us. Machines, on the other hand, are still learning how to apprehend various human activities, and we are teaching them based on our knowledge and understandings of the task. Considering the fact that machines (or computers) were nothing but simple calculators to solve arithmetic problems just sixty years ago [1], their understanding of complex concepts has come a long way. Machine Learning (ML), as a part of the Artificial Intelligence (AI), has given machines the capacity to interpret various situations in their surroundings and respond accordingly like humans. Human Activity Recognition (HAR) is being researched since the early 1980s because of its promise in many applied areas. However, the significant breakthroughs in this field have come within the last two decades [2]. The recent developments in microelectronics, sensor technology, and computer systems have made it possible to collect more fundamental information from human movements, and the advanced ML techniques have made that information more comprehensible to the machines.

There are several approaches to collect HAR data from the participating subjects; broadly, they fall into one of the two categories – namely camera-based recording or sensor-based recording [3]. In the former approach, one or more video cameras are set up to record the activities of a subject for a certain amount of time, and then the recognition is performed using video analysis and processing techniques. The later one utilizes various types of sensors to track the movements of the subject. This approach can be further classified based on the type of sensors used – whether they involve wearable body sensors or the external ones [2]. External sensors are placed in predetermined points of interest on the subjects' body, whereas wearable sensors require to be attached to the subject while collecting data. Each of these techniques has its advantages, shortcomings, and apposite applications. Some recognition techniques even combine multiple recording techniques to collect more relevant data and make the corresponding actions more interpretable to the machines. The applications of HAR include intelligent surveillance, haptics, human-computer interaction, motion or gesture-controlled devices, automatic health-care monitoring systems, prosthetics, and robotics. Despite many advancements, HAR is still a challenging task because of the articulated nature of human activities, the involvement of external objects in human interactions, and complicated spatiotemporal structures of the action signals [4]. Success in recognizing these activities requires advanced signal and image processing techniques, as well as sophisticated ML algorithms. Since the absolute performance is yet to be achieved, HAR remains a tending field to the researchers.

HAR is a practical field that requires the knowledge of both Biomedical Engineering and Computer Science. Because of its realistic nature, the machines need real-life human action data to learn from. Various Universities and laborites around the world provide datasets that contain information on various human motions in the form of analog signals. Amongst the most renowned HAR datasets, one is hosted by the University of California Irvine (UCI) in their Machine Learning Repository, which is commonly known as the UCI HAR dataset [5]. Reference [6] is the first work on this human activity dataset, where the authors achieved an 89.35% classification accuracy employing a hardware-friendly version of the Support Vector Machine (HF-SVM). However, the paper was published before the dataset was publicized in December 2012. In 2013, the same group of authors reported a 7% improvement in the classification accuracy by mapping 561 statistical features from the raw data [7]. The paper also outlines the data collection procedure, detailed system architecture, data specification, and information on data processing. In the same year, the authors published another article focusing on the energy efficiency of the model where they extracted a different set of features and classified them using multiple HF-SVMs with different Look-Up-Tables (LUTs) [8]. As of 2019, numerous studies have been conducted on the UCI HAR dataset by numerous groups of researchers which have resulted in a number of separate methods employing a wide range of feature extraction, feature selection and ML techniques. Reference [9], where the authors experimented with various versions on the neural network to classify HAR signals, and [10], where an approach based on Semi-Supervised Active Learning (SSAL) was described are some of the most recent works incorporating the UCI HAR dataset. Among all the studies conducted on the cited dataset, [11] and [12] provide the highest accuracy. In this study, we are going to describe a classification model based on a multilayer Convolutional Neural Network (CNN) to classify different human activities. Instead of the statistical features described in [7] and [8], we extracted frequency and power information from the signals and fed them to a multichannel CNN model. The outputs were concatenated prior to the final classification. Necessary figures, flowcharts, and tables have been provided after each step of the procedure to simplify the narration and support the methodology.

The rest of the paper is organized as follows. Section II elaborates the methodology of the study along with brief discussions on the UCI HAR dataset, formation of the operational dataset and the basics of CNN. Section III presents the obtained results and evaluates the method based on some well-known parameters. Finally, Section IV provides a summary and expounds some scopes for future research.

## II. METHODOLOGY

This study aims to classify the HAR signals of the UCI HAR dataset employing a two-channel CNN model, as shown in Fig. 1. Like all the supervised ML techniques, this algorithm has two stages (i.e., training stage and testing stage). The training stage requires a set of data samples containing various attributes measured from subjects while performing various predefined activities. The supervised learning technique then try to make some "sense" out of the data, find out how the samples that belong to the same class are similar to each other while samples from different classes are diverse, then builds one or more internal models focusing on the crucial attributes that can highlight those contrasting properties to carry out the classification [2]. However, merely feeding the raw data collected from the sensors into the classifier might not be a good idea, because more often than not these time-domain signals contain noise, interference, missing values, and most importantly, time-domain attributes are simply not good enough to make the distinguishable properties perceptible to the classifiers. That is why researchers spend so much time finding and selecting meaningful features from various types of real-life time-varying signals, which is also known as feature engineering [13]. Now, although paper [7] and [8] worked with that statistical features and acquired decent results, in this study, we are taking on a slightly different approach. We are extracting frequency and power information (or features) from the raw time-domain accelerometer signals and then feeding these two sets of samples into a two-channel CNN. In the training stage, a preordained portion of the dataset is used to train the machine and build a feasible model, which is then evaluated over the remaining samples.

*A. UCI HAR Using Smartphones Dataset*

The UCI HAR dataset contains mobility information that was collected from 30 people of different ages (ranging from 19 to 48 years), genders, heights and weights using a wrist-mounted smartphone (Samsung Galaxy S II). The smartphone has integrated accelerometer and gyroscope. Action data was recorded using these sensors while each of the subjects was performing six predefined tasks, which, according to the jargon of ML, represent six different classes. Three-axial linear acceleration and three-axial angular velocity data were acquired at a steady rate of 50 Hz [7]. The collected samples were labeled manually afterward. Before putting in the dataset, the samples were pre-processed using a median filter for noise cancellation and a third-order low-pass Butterworth filter having a 20 Hz cutoff frequency. The available dataset contains 10,299 samples which are separated into two sets (i.e., a training set and a test set). The former one contains 7,352 samples (71.39%), whereas the latter one is comprised of the rest 2,947 samples (28.61%). Table I provides more details about the contents of the dataset along with the class identifications and their labels.

TABLE I. ACTIVITY CLASSES, THEIR LABELS AND SAMPLE RATIOS

| Activity | Class Label | Training Samples | Test Samples |
|---|---|---|---|
| 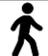 | Wlk | 1226 (16.67%) | 496 (16.83%) |
| 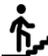 | WUp | 1073 (14.60%) | 471 (15.99%) |
| 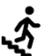 | WDn | 986 (13.41%) | 420 (14.25%) |
| 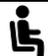 | Sit | 1286 (17.49%) | 491 (16.66%) |
| 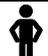 | Stn | 1374 (18.69%) | 532 (18.05%) |
| 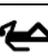 | Lay | 1407 (19.14%) | 537 (18.22%) |

Fig. 2 provides the t-distributed stochastic neighbor embedding (t-SNE) graph of the operational dataset, which illustrates how the samples of different classes are distributed over a two-dimensional plane. We can see that despite belonging to separate classes, samples of four different classes are all cramped together, which is not ideal for classification.

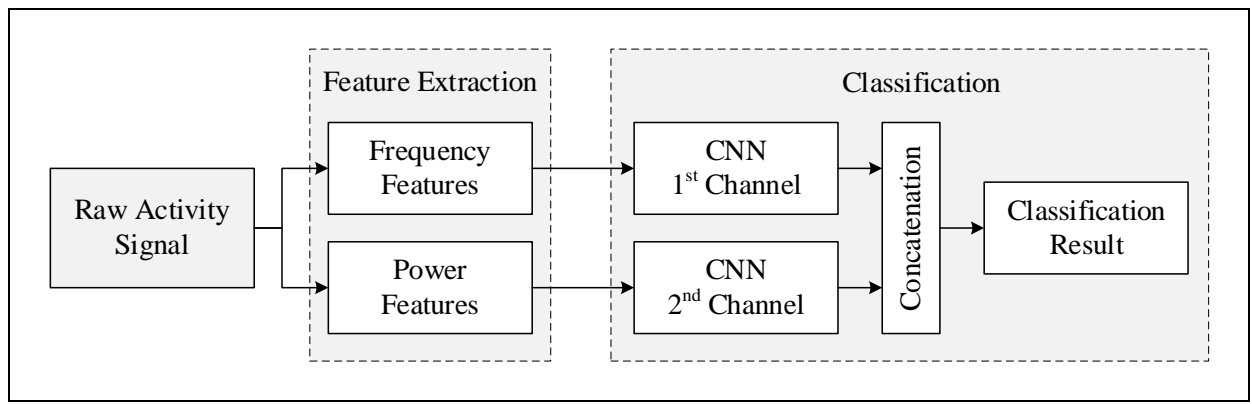

Fig. 1. Proposed CNN-based HAR classification model.

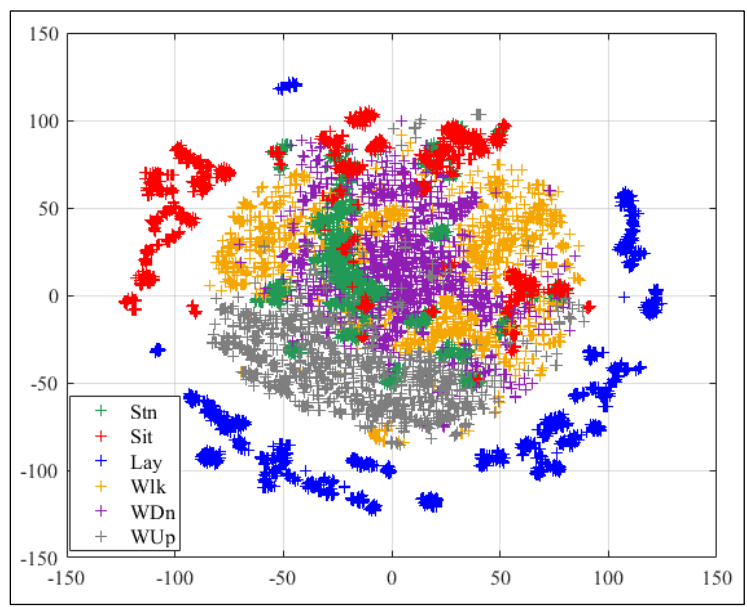

Fig. 2. t-SNE of the operational HAR samples.

## B. Frequency Features

From the perspective of the frequency contents, a Human action signal can be viewed as a comprised form of multiple sinusoidal signals of different frequencies. The frequency information of the human action signal simply refers to the values of those frequencies and the amplitudes of those signals in those frequencies. Digital Signal Processing (DSP) offers multiple methods to extract this information to form a HAR signal. In this study, we are going to use the well-known Fast Fourier Transform (FFT). The algorithm of FFT was developed by James Cooley and John Tukey in 1965 as a faster version of the then-popular Discrete Cosine Transform (DFT) to calculate the frequency components of a time-domain signal [14]. If we consider a $N$-point time-series signal $x(N)$, its $N$-point DFT is defined by:

$$X^{(N)}(k) = \sum_{n=0}^{N-1} x_n e^{\frac{-i2\pi kn}{N}} \quad (1)$$

where $k = 0,1,2,...,N-1$. FFT is an algorithm for computing the $N$-point DFT with a computational complexity of $\mathcal{O}(N \log N)$. Fig. 3 presents the t-SNE graph of the HAR signals after extracting their frequency features using FFT. It is noticeable that in this figure, the samples that belong to the same class are more clustered, and samples of different classes are more disjoint than Fig. 2 This means that the classifier will find these samples easier to classify.

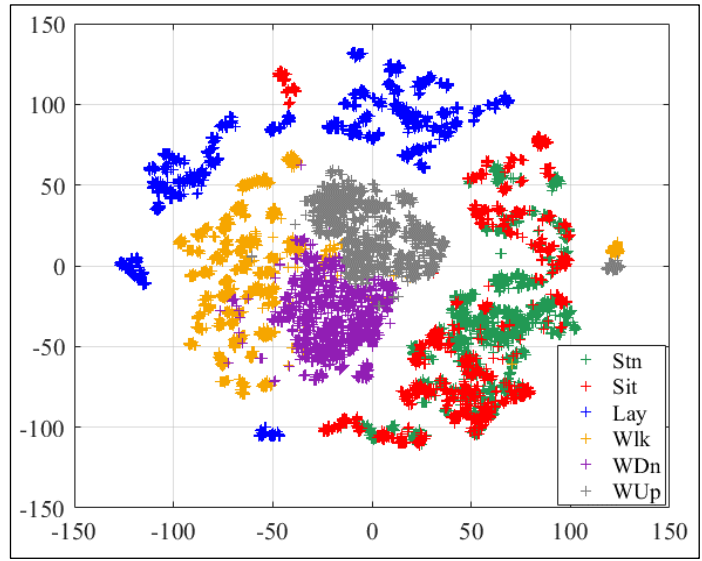

Fig. 3. t-SNE of the HAR samples based on their frequency information.

## C. Power Features

The power spectrum of a signal simply represents the distribution of power into various frequency components present in it. There are multiple methods that can extract the power information of a signal. The pwelch algorithm gives us the Welch's power spectral density (PSD) estimate of the input signal calculated using Welch's overlapped segment averaging estimator. The signal processing technique was introduced by P.D. Welch in 1967 [15]. Two distinguishable properties of this algorithm are – first, each data segments are allowed to overlap, and second, each data segment is windowed before computing the periodogram. According to the pwelch algorithm, for $N$ total observations and $L$ subsamples, the windowed periodogram of the $i$th data segment can be defined as:

$$\widehat{\Phi}_i(\omega) = \frac{1}{OP} \left| \sum_{t=1}^{O} u(t) y_i(t) e^{-j\omega t} \right|^2 \quad (2)$$

where $O$ is the number of observations in each subsample, $u(t)$ is the temporal window, $y_i(t)$ is the $i$th data segment. Now, the power $P$ of the temporal window $\{u(t)\}$ can be written as,

$$P = \frac{1}{M} \sum_{t=1}^{O} |u(t)|^2 \quad (3)$$

If we calculate the average of windowed periodograms denoted in (2), we will get the final Welch estimate of PSD (pwelch):

$$\widehat{\Phi}_W(\omega) = \frac{1}{L} \sum_{i=1}^{L} \widehat{\Phi}_i(\omega) \quad (4)$$

Overlapping between two successive data segments allows us to achieve more periodograms to be averaged in (4), which decreases the variance of the estimation. Windowing, on the other hand, allows us to get more control over the resolution properties of the estimation [15]. Moreover, the temporal window $\{u(t)\}$ can make the sequent subsamples less correlated to each other, even after being overlapped. Fig. 4 presents the t-SNE graph of the HAR signals after extracting their power features.

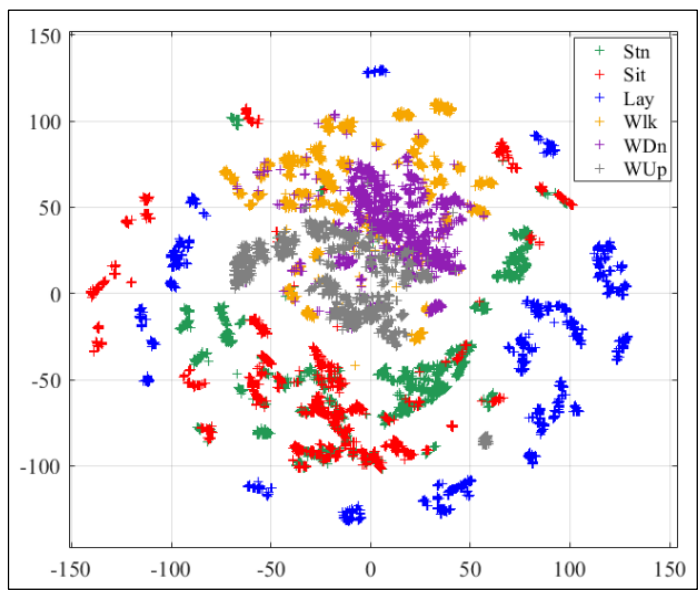

Fig. 4. t-SNE of the HAR samples based on their power information.

Close examination of Fig. 3 and Fig. 4 reveals that the samples are distinguishable based on either feature set, yet in both cases, there are some areas where samples of different classes overlap. Using both sets of information would be more useful to identify the discernible properties of the samples.

*D. Operational Dataset Preparation*

Each of the samples of the UCI HAR dataset contains a subject's body acceleration (Body_acc), Triaxial Angular velocity (Body_gyro) and total acceleration (Total_acc) data in three axes (namely X, Y, and Z) while performing an assigned activity. Fig. 5 shows how each set of data was processed individually to extract frequency and power information from them. After that, the frequency and power features of each signal were concatenated to form a complete feature set that represents the corresponding HAR signal in the classification stage. Finally, the label of the associated class is inserted at the terminal point of the signal.

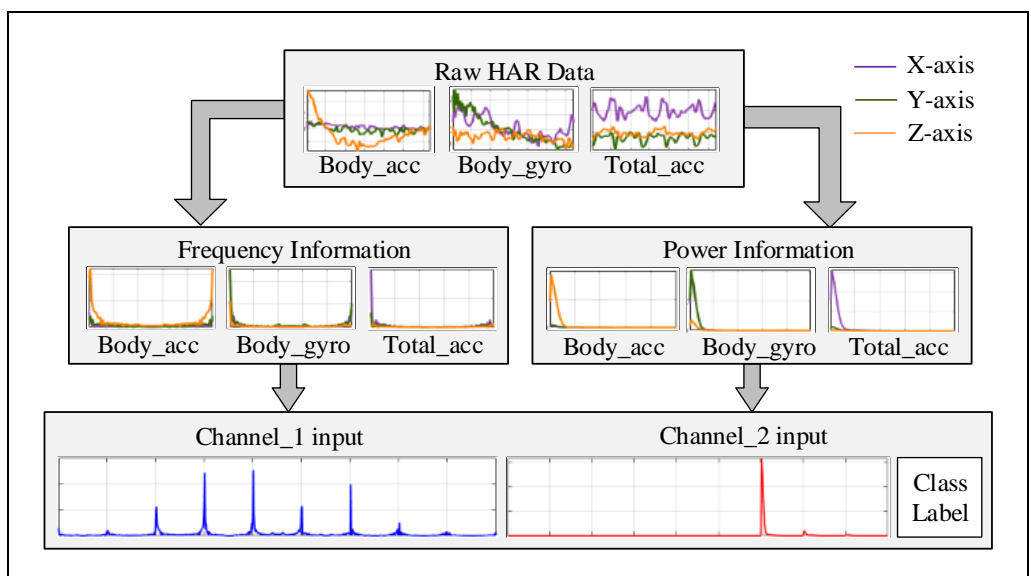

Fig. 5. The feature set corresponding to a sample signal for classification.

*E. Multichannel CNN for HAR*

CNNs are biologically inspired networks for processing data that has a known, grid-like topology. CNN works based on convolution, which is a mathematical operation that slides one function over another and measures the integral of their point-wise multiplication [16]. Convolutional networks are neural networks that use convolution in lieu of general matrix multiplication in at least one of their layers [17]. The idea behind the CNN was obtained from Hubel and Wiesel's interpretation of the operation of the cat's visual cortex, where they observed that specific portions of the visual field excite particular neurons [18]. Although CCN was developed to work mainly with the images that have high dimensionality, they are equally effective on various types of analog and digital signals. CNN has been extensively used in speech recognition, audio processing, machine fault classification, and various types of biomedical signal recognition such as Electrocardiogram (ECG), electroencephalogram (EEG), Electromyography (EMG). As stated before, we are incorporating two CNNs (specified as channels) in our classification model; one of them will process the frequency features, and the other will work with the power features as shown in Fig. 6. The basic parameters for both the channels are the same.

To explain how each of the CNN units works, let us consider a sequence of $M$-dimensional observations arranged in a mini-batch of length $L$, the output of a hidden layer $h$ at node $n$ and time $t$ is calculated from the input $x_t \in \mathcal{R}^M$ as:

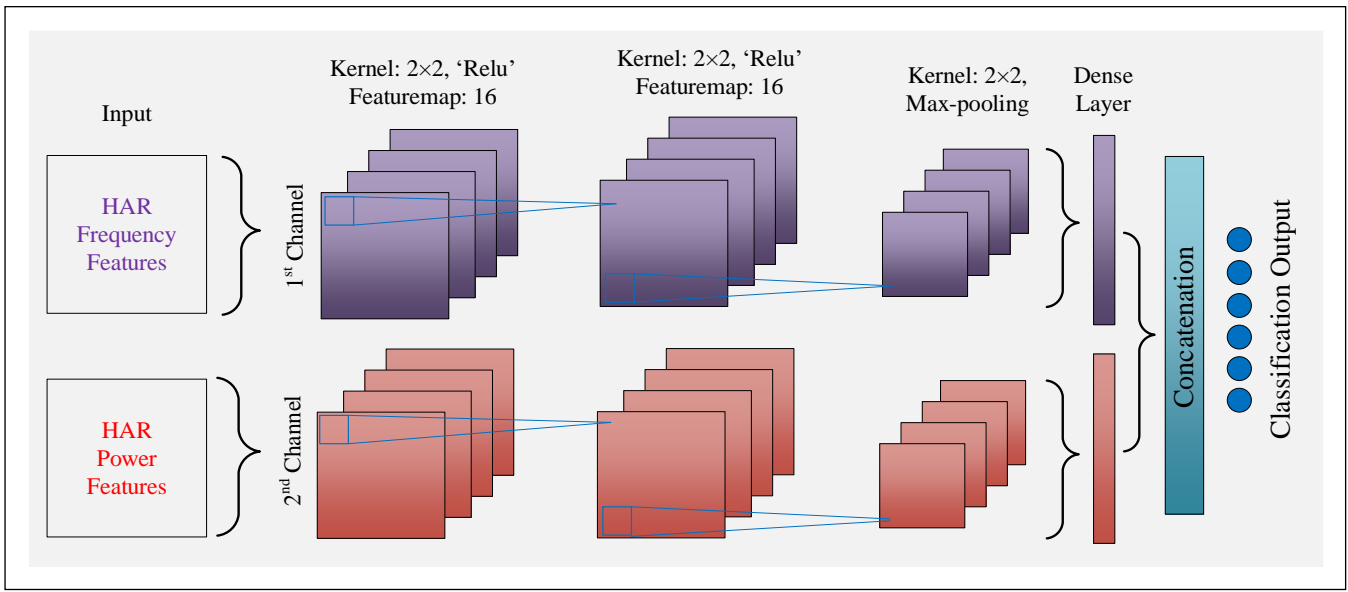

Fig. 6. Proposed two-channel CNN for HAR classification

$$h_{n,t} = \sigma \left( \sum_{i=0}^{M-1} \omega_{n,i} x_{i,t} + b_n \right) \quad (5)$$

where $0 \le t < L$, $\sigma$ is a non-linear function (such as sigmoid), $\omega_{n,i}$ are the weights, $b_n$ is the bias [19]. The weight and bias depend on the identity of the neuron $n$. Usually, a convolutional layer operating on both the time and frequency dimensions arranges the weights as a two-dimensional patch whose parameters also depend on $n$. We can evaluate the output of such a layer at different points of the matrix $(s, t)$:

$$h_{n,s,t} = \sigma \left[ \sum_{i=s}^{s+m-1} \left( \sum_{j=t}^{t+m'-1} \omega_{n,i-s,j-t} x_{i,j} + b_n \right) \right] \quad (6)$$

where the weight patch $\omega \in \mathcal{R}^{m \times m'} (m < M, m' < L)$ is multiplied with a part of the input that covers the direct neighborhood of the position $(s, t)$. The set of outputs of a convolutional unit $n$ is called the *feature map* which are similar to a feature stream extracted by $n$. For the sake of simplicity, we are suppressing the details on patch symmetry and boundary handling. However, within this framework of training, it is convenient to have a feature map that has the same number of columns as the mini-batch [19].

Now, in the CNN architecture, the output of each filter of the first layer is considered the input of the following next layer. If we have $R$ input streams that are arranged in mini-batches of the same dimension, we can calculate the corresponding output using (6):

$$h_{n,s,t} = \sigma \left[ \sum_{i=s}^{s+m-1} \left\{ \sum_{j=t}^{t+m'-1} \left( \sum_{r=0}^{R-1} \omega_{n,i-s,j-t,r} x_{i,j,r} + b_n \right) \right\} \right] \quad (7)$$

In the case of HAR features, the output of a convolutional unit $n$ can be defined as:

$$h_{n,t} = \sigma \left( \sum_{i=s}^{p+m-1} \omega_{n,i-s} x_i + b_n \right) \quad (8)$$

where $\omega \in \mathcal{R}^m$ is a weight vector and $p$ is the position within the output vector. The step size $z$ is chosen such that the output feature stream is only calculated for positions specified by $\{h_{n,s\cdot z} : 0 \leq s \cdot z < M - m\}$. The dense layer outputs of the CNN channels are then concatenated to acquire the final classification output.

## III. RESULTS AND DISCUSSION

In the previous sections, we have discussed the contents of the UCI HAR dataset, our approach to classifying the samples of six different classes contained in it, as well as the techniques and methods that we have employed in the proposed methodology. In this section, we will present the findings of the study. Following the described procedure, we set a classification model where the provided training samples were used to train the two-channel CNN model, and the rest of the samples were used to test it. The result yields a classification accuracy of 95.25% on the test samples. Fig. 7 presents the classification accuracies on both the training and testing samples at each epoch. As seen in the figure, the training accuracy gradually increased with each epoch. The test accuracy, on the other hand, kept fluctuating around the 94% mark and peaked at the 38th and the final epoch. The performance of the model was slightly unstable throughout the first 20 epochs, but it became pretty stable afterward.

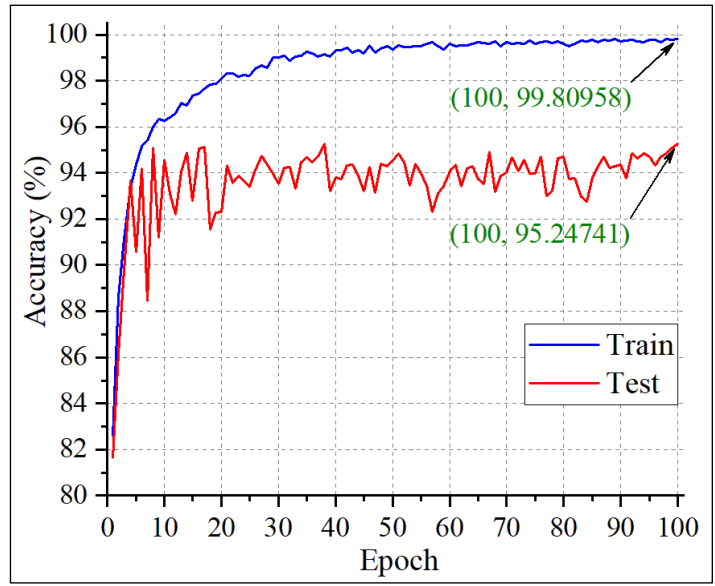

Fig. 7. Train and test classification accuracies at each epoch.

Apart from the classification accuracy, the confusion matrix of classification is another useful tool to judge the performance of the classification model [20]. The confusion matrix provides more details on the output of the classification process. It tells us how many samples of each class were tested and how many of them were classified correctly based on the model. In the best-case scenario (i.e., when the classification accuracy would be 100%) only the diagonal boxes of the matrix would contain non-zero values or the number of tested samples of the corresponding class, and all the other boxes would contain zeros. Fig. 8 provides the confusion matrix of the epoch of our model for HAR classification. It is apparent that the model works very well while distinguishing five of the six classes (*Walking, Walking-Upstairs, Walking-Downstairs, and Laying*) registering over 95% individual classification accuracies for each class. However, the model faces some difficulties while differentiating the *Sitting* states from the *Standing* states, as we can see that 12.8% samples of the former class have been misclassified as the later one. The performance can also improve while separating samples that belong to the three different classes of Walking.

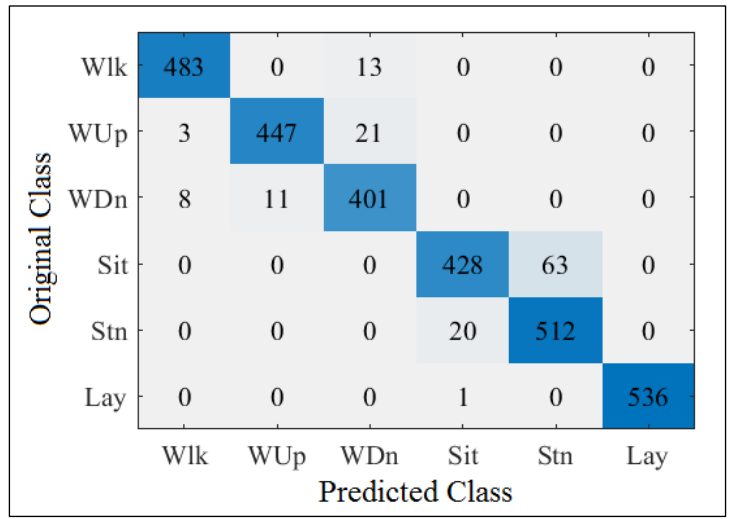

Fig. 8. Confusion matrix of the HAR classification.

Precision, recall, and F1-score are three commonly used parameters to measure the potency of a classifier. The precision of a class refers to the rate of the correctly classified samples within the positively classified samples of that class [21]. High precision indicates a low false-positive rate and vice versa. Recall of a class, on the other hand, refers to the fraction of the truly positive instances of the class that the classifier recognizes. A high recall indicates that the classifier made a handful of false-negative predictions and vice versa. Usually, the precision and recall values of all the classes are averaged and expressed as the precision and recall of the model, respectively. F1-score is simply the harmonic mean or the weighted average of the precision and recall values. Fig. 9 depicts the precision, recall, and F1-scores at each epoch of the described CNN-based classification model. All three matrices have very close values for both the train and test classifications, which is why they have been represented with only two lines.

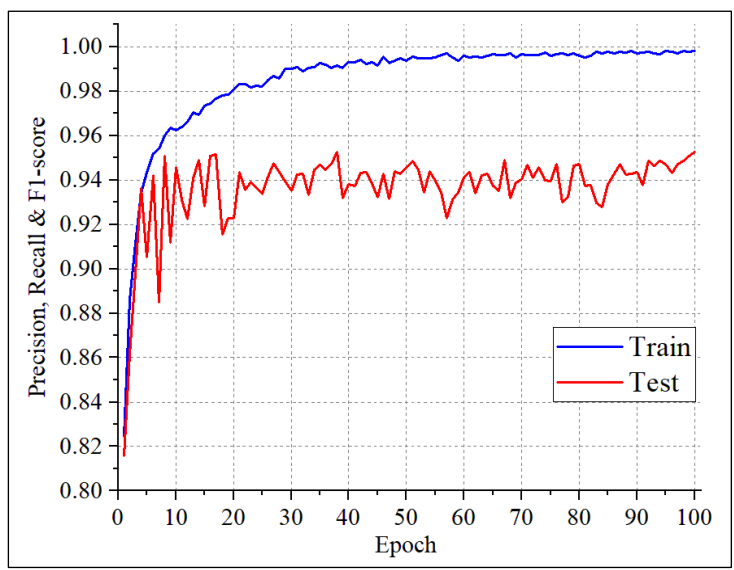

Fig. 9. Precision, recall, and F1-scores at each epoch.

Furthermore, to clarify the performance measure of the model, we present the Receiver Operating Characteristics (ROC) curves of each class of the classification in Fig. 10. The ROC curve is not affected by the unbalanced proportions of class instances presented in the dataset and provides a more accurate measure of the classification [21]. It plots the true positive rate (recall) of the classifier against the false positive rate (fall-out). The ROC curve of a classifier that predicts its classes inaccurately or randomly would follow the diagonal dashed line of the graph. The more accurate the classifier is, the more distance it would keep form the dashed line. A perfect classifier would have of all curves touching the top-left corner. However, since our classification outcome was less than ideal, the curves went very close to the corner but did not make contact, except for the *Laying* curve which has the highest individual classification accuracy (99.81%).

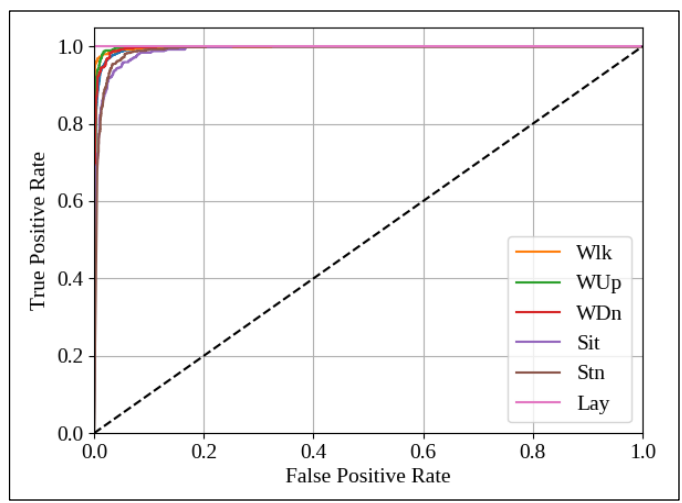

Fig. 10. The ROC-AUC curve of the classification.

To put the outcome of our classification model in context, we have compared our acquired results with that of other four similar studies involving the UCI HAR dataset in Table II. The table shows that in terms of classification accuracy the proposed method outperforms the methods described in [6], [9] and [10] by 5.9%, 0.17%, and 14.01% respectively. Only [7] and [12] have attained better performances than the proposed model. However, [6] and [9] were more successful in classifying the samples of the *Sitting* class than [7], and our model has higher individual accuracy in some classes.

TABLE II. COMPARISON WITH SIMILAR METHODS

| | Method | Our | [6] | [7] | [9] | [10] | [12] |
|---|---|---|---|---|---|---|---|
| Accuracy (%) | Wlk | 97.38 | 95.61 | 99.19 | 97.78 | 84.84 | 97.8 |
| | WUp | 94.90 | 69.85 | 95.75 | 92.56 | 80.83 | 97.7 |
| | WDn | 95.48 | 83.22 | 97.62 | 98.75 | 80.80 | 98.1 |
| | Sit | 87.17 | 92.96 | 87.98 | 96.77 | 79.46 | 98.9 |
| | Stn | 96.24 | 96.43 | 97.37 | 87.08 | 80.69 | 92.5 |
| | Lay | 99.81 | 100 | 100 | 99.81 | 80.86 | 100 |
| | **Total** | **95.25** | **89.35** | **96.37** | **95.08** | **81.24** | **97.5** |
| Precision (%) | | 95.32 | 89.93 | 96.58 | 94.99 | 81.66 | 97.3 |
| Recall (%) | | 95.16 | 89.68 | 96.32 | 95.46 | 81.25 | 97.5 |
| F1-score (%) | | 95.24 | 89.8 | 96.45 | 95.22 | 81.45 | 97.4 |

## IV. CONCLUSIONS

The paper describes a multichannel CNN-based HAR classification model and tests it on the UCI HAR dataset extracting the frequency and power features of the samples. The obtained results yield a 95.25% classification accuracy. However, the model can be further modified by tuning specific parameters of CNN and adding more nodes and layers in the CNN architecture. A new set of features can also be extracted and fed in an additional channel of CNN to improve the model's performance, which is subjected to future studies. The issue with the low classification accuracy of the *Sitting* class must be addressed as well. We are also interested in evaluating our model using other HAR datasets, including an updated version of the UCI HAR dataset that contains Postural Transitions [22].